# Antenna Optimization Using a New Evolutionary Algorithm Based on Tukey-Lambda Probability Distribution

Vahraz Jamnejad [1], Ahmad Hoorfar [2]

*1. Jet Propulsion Laboratory, California Institute of Technology, Pasadena, CA 19109, USA*

[1] vahraz.jamnejad@jpl.nasa.gov

*2. Department of Electrical and Computer Engineering, Villanova University, Villanova, PA, 19085, USA*

[2] ahoorfar@villanova.edu

*Abstract*—In this paper, we introduce a new evolutionary optimization algorithm based on Tukey's symmetric lambda distribution. Tukey distribution is defined by 3 parameters, the shape parameter, the scale parameter, and the location parameter or average value. Various other distributions can be approximated by changing the shape parameter, and as a result can encompass a large class of probability distributions. In addition, Because of these attributes, an Evolutionary Programming (EP) algorithm with Tukey mutation operator may perform well in a large class of optimization problems. Various schemes in implementation of EP with Tukey distribution are discussed, and the resulting algorithms are applied to selected test functions and antenna design problems.

## 1. INTRODUCTION

The Evolutionary Programming (EP) [1-2] is a nature-based stochastic global optimization technique that can directly work with continuous parameters and unlike the Genetic Algorithms (GAs), which use a variation operator composed of Cross-over and mutation, utilizes a mutation only variation operator where adaptive and/or self-adaptive techniques exist, or can easily be designed, for adapting the parameters of mutation operator during the evolution process.. Conventional implementation of EP for continuous parameter optimization uses Gaussian mutations. An implementation of EP with Cauchy mutation operator, however, has empirically been shown to outperform EP using the Gaussian mutations for optimizations of multi-modal functions with many local optima. The first implementation of EP for electromagnetic and antenna applications was given in [3], and an extensive review of various forms of EP using Gaussian, Cauchy and a hybrid linear combination of these operators in optimization of antenna and microwave structures is reported in [4]. In addition, a comparative study of EP, GAs and Particle Swarm Optimization (PSO) was given in [5] where it was shown that EP outperforms other algorithms in several standard antenna optimization problems. Also, application of EP to complex corrugated horn antenna design with a large number of optimization parameters was given in [6].

Furthermore in [7], we introduced a new adaptive scheme for an EP algorithm using a mutation operator based on Lévy probability distribution and compared its performance in antenna optimization to other adaptive and non-adaptive schemes reported in [8],

Lévy distribution has a parameter $\alpha$ that controls the shape of the distribution; in the limits when $\alpha$ approaches values of 1 and 2, the Lévy distribution reduces to Cauchy and Gaussian distributions, respectively. As a result, an adaptive selection of the shape-parameter can be used to exploit the desirable properties of these two distributions.

An alternative parametric probability distribution that can include Cauchy and Gaussian distributions as special cases is that of Tukey's symmetric lambda distribution [9-11]. In general, Tukey distribution is defined by 3 parameters, the shape parameter, $\lambda$, the scale parameter, $\beta$, and the location parameter or average value, $\alpha$. Various other distributions can be approximated by changing the shape parameter, and as a result can encompass a large class of probability distributions. In addition, Unlike Lévy distribution, it has its inverse of the cumulative distribution function in closed form, which facilitates the generation of its corresponding random variables. These attributes suggest that an EP algorithm with Tukey mutation operator may perform well in a large class of optimization problems.

In this paper, we first investigate the implementation and performance of EP with Tukey mutation in optimization of selected test functions. We then apply this new EP technique to selected antenna design problems. An abstract of this work was appeared in [15].

## 2. EP WITH TUKEY-LAMBDA MUTATION

The standard EP algorithm with self-adaptive mutation operator for optimization of an n-dimensional objective function $\phi(\bar{x})$, $\bar{x} = [x(1), x(2), \ldots, x(n)]$ consists of five basic steps: initialization, fitness evaluation, mutation, tournament and selection. Here we concentrate on the mutation step. First, let us assume an initial population of µ individuals is formed through a uniform random or a biased distribution. In EP with Gaussian mutation operator (GMO), each individual is taken as a pair of real-valued vectors, $(\bar{x}_i, \bar{\eta}_i)$, $\forall i \in \{1,...\mu\}$, where each parent $(\bar{x}_i, \bar{\eta}_i)$ creates a single offspring $(\bar{x}_i', \bar{\eta}_i')$ by:

$$\eta_i'(j) = \eta_i(j) e^{[\tau' N(0,I) + \tau N_j(0,I)]} \quad (1.1)$$
$$x_i'(j) = x_i(j) + \eta_i(j) N_j(0,1) \quad (1.2)$$

for j = 0,1,2,....n, where *x(j) and η(j)* and are the jth components of the solution vector and the variance vector, respectively. N(0,1) denotes a one-dimensional random variable with a Gaussian distribution of mean zero and standard deviation one. Nj(0,1) indicates that the random variable is generated anew for each value of j. The scale factors τ and τ' are commonly set to $\left(\sqrt{2\sqrt{n}}\right)^{-1}$ and $\left(\sqrt{2n}\right)^{-1}$, respectively, where n is the dimension of the search space. In EP with Cauchy mutation operator (CMO), the offsprings are still generated according to (1), but with a Cauchy mutation replacing the Gaussian mutation in the second equation.

In our proposed EP with Tukey mutation operator (TMO), the mutation step is the same as above but now the equation in (1.2) takes the form:

$$x_i'(j) = x_i'(j) + \eta_i'(j) T_j(\beta, \lambda) \quad (2)$$

where $T_j(\beta,\lambda)$ indicates a random variable generated anew for each value of *j* from the Tukey-Lambda probability distribution given by,

$$F(x), \quad x \in R \quad (3)$$

The inverse of (3), which represents the random variable, $T(\beta,\lambda)$ is given in closed-form as [10]:

$$x = F^{-1}(p) = \alpha + \beta \frac{[p^\lambda - (1-p)^\lambda]}{\lambda} \quad (4)$$

The above distribution is defined by the shape parameter, $\lambda$, the scale parameter, $\beta$, and the location parameter, or average value, $\alpha$.

For a given value of $\alpha$, different combination of $\lambda$ and $\beta$, result in different shapes of the corresponding Tukey-Lambda probability distribution. We have investigated three different schemes in selection of these parameters as applied to the EP steps in (4).

*Scheme 1*. In this scheme, we use a population of µ = k and randomly select $\beta$ and $\lambda$ in (4) as:

$$\beta_i = a_{1i} \text{rand}(1,3) + a_{2i} \text{rand}(1,i)$$
$$\lambda_i = b_{1i} \text{rand}(1,i)) + b_{2i} \text{rand}(1,i) \quad ; i=1, 2, 3$$

where rand (1) is a random number with uniform distribution in the interval [0, 1], and $a_1$, $a_2$ and $b_1$, $b_2$ are constants that are used to put constraints on the ranges of β and λ.

*Scheme 2*. In this scheme, we use a population of µ = 3k, where for each $k_i$ sub-population we randomly select β and λ similar to the above procedure in (5), i.e.,

$$\beta_i = a_{1i} \text{rand}(1,3) + a_{2i} \text{rand}(1,i)$$
$$\lambda_i = b_{1i} \text{rand}(1,i)) + b_{2i} \text{rand}(1,i) \quad ; i=1, 2, 3$$

*Scheme 3*. In this scheme, we also use a population of µ = 3k, where for the first two sub-populations, k=1 and k=2, we select $\beta_1$, $\lambda_1$ and $\beta_2$, $\lambda_2$ according to their corresponding values for which Tukey mutation operator reduces to the standard Gaussian and Cauchy mutations. The parameters of the last sub-population in this scheme are randomly chosen according to the above procedure in Scheme 1. This third scheme, is basically equivalent to generating offsprings according to:

$$x_i'(j) = x_i'(j) + \eta_i'(j) N_j(0,1) \quad ; i = [k+1, 2k] \quad (5)$$
$$x_i'(j) = x_i'(j) + \eta_i'(j) C_j(0,1) \quad ; i = [2k+1, 3k] \quad (6)$$
$$x_i'(j) = x_i'(j) + \eta_i'(j) T_j(\beta_3, \lambda_3) \quad ; i = [3k+1, 4k] \quad (7)$$

## 3. EXAMPLES OF FUNCTION OPTIMIZATION

In order to compare the convergence rate of the three selection schemes in the EP-Tukey algorithm described in the previous section, we have applied those alternative schemes to two commonly used test functions: 20-D Ackley function and 2-D Rosenbrook functions [2]. In order to make the total number of the fitness evaluation fixed, the population sizes of 120, 60 and 60 were used in Schemes 1, 2 and 3, respectively. In each case the optimization was applied to 25 trials. Figures 1-4 show the fitness trajectories of the overall best solution and the average best solution population member averaged over 25 trials. As can be seen for both functions, Scheme 2 provides the best solution when averaged over 25 trials. We note, however that Scheme 3 results in the overall best solution, particularly for the Ackley function, which has an extremely large number of local minima. This can be attributed to the fact that Scheme 3 forces 1/3 of the population members to go through the Cauchy mutation, which has been shown to be particularly effective in unconstrained optimization of objective functions with many local minima.

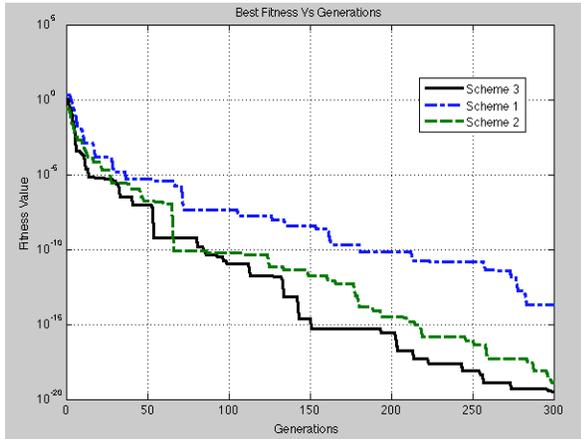

Figure 1: Convergence rate of best solution in optimization of Rosenbrook function

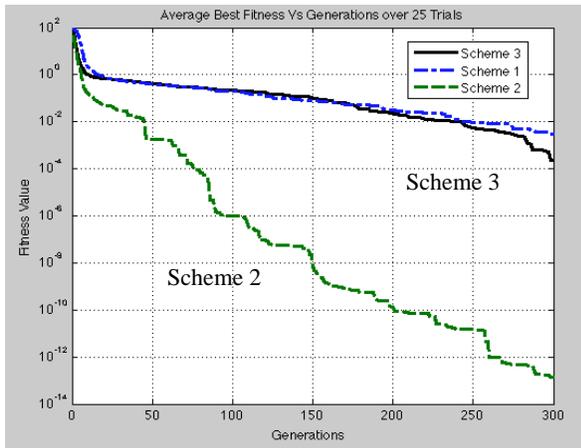

Figure 2: Fitness trajectory of average best solution in optimization of Rosenbrook function

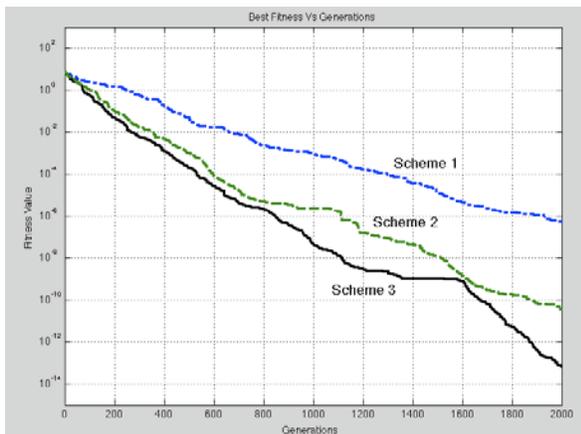

Figure 3: Convergence rate of best solution in optimization of Ackley function

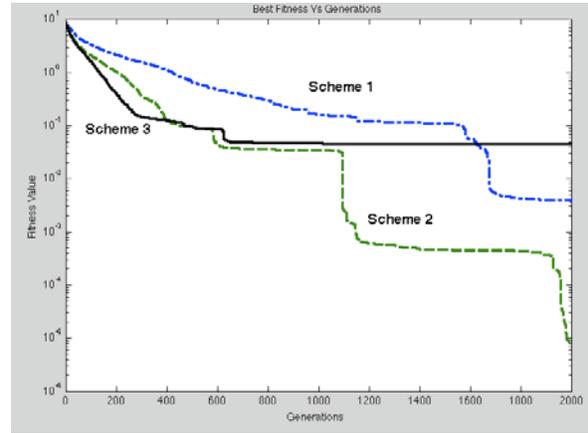

Figure 4: Fitness trajectory of average best solution in optimization of Ackley function

## 4. EXAMPLE OF ANTENNA OPTIMIZATION

The standard implementation of EP algorithm with Gaussian mutation has been previously used in optimization of corrugated horn antennas for space applications [6, 12]. Here, as an example of this new EP-inspired optimization technique, we apply it to the design of the classic Dragonian antenna system [13-14]. The Dragonian dual-reflector antenna system employs a paraboloidal main reflector illuminated by a concave hyperboloidal subreflector, and has very good performance characteristics. The relatively large offset distance and focal length of the main reflector, and the avoidance of caustics between the two reflector surfaces, which is a consequence of the concave nature of the subreflector, result in relatively flat reflectors with a wide field of view capability. In addition, the Dragonian antennas can be designed to be efficiently illuminated by electrically small feeds. In general, any dual reflector system is completely specified by 5 parameters (see Fig. 5, adapted from [14]). The five parameters that we choose are:

$D$, diameter of the main reflector,
$\theta_e$, half-cone angle of feed pattern incident on the reflector,
$\theta_p$, angle of feed central axis with axis of the main reflector, z,
$\theta_0$, angle of central ray hitting the reflector with main axis, z, (main reflector offset angle)
$f_{12}=2c$, Inter-focal distance of the hyperbolic subreflector.

We consider a particular design of the antenna for minimum spillover and zero geometrical-optics cross polarization [14]. The condition for this design is given by:

$$\tan(\frac{\beta}{2}) = M \tan\left(\frac{\theta_p - \beta}{2}\right) \quad (8)$$

In which M is the magnification factor, related to eccentricity, $M = (e+1)/(e-1)$, of the hyperboloid. This, together with the definition of the angular magnification relation of the hyperboloid,

$$\tan\left(\frac{\theta_p - \beta}{2}\right) = M \tan(\frac{\beta - \theta_0}{2}) \quad (9)$$

can be used to obtain,

$$\cot\left(\beta - \frac{\theta_p}{2}\right) = 2\cot(\frac{\theta_0}{2}) - \cot(\frac{\theta_p}{2}) \quad (10)$$

Thus, given the parameters $\theta_p$, $\theta_0$, we find the angle of feed pattern axis with the subreflector axis, $\alpha$, the tilt angle of subreflector hyperboloid axis with respect to the main reflector axis, $\beta$, and the magnification factor, $M$, of the hyperboloid, after some manipulation, as follows

$$\alpha = \theta_p/2 - \gamma = \theta_p/2 - \cot^{-1}[2\cot(\frac{\theta_0}{2}) - \cot(\frac{\theta_p}{2})]$$

$$\beta = \theta_p/2 + \gamma = \theta_p/2 + \cot^{-1}[2\cot(\frac{\theta_0}{2}) - \cot(\frac{\theta_p}{2})]$$

$$M = \frac{\tan(\beta/2)}{\tan(\alpha/2)} \quad (11)$$

Then other relevant parameters are obtained as
Main reflector focal length, $F = F_e \frac{\sin(\beta)}{\sin(\alpha)}$

With equivalent focal length, $F_e = \frac{D}{4\tan(\theta_e/2)}$

Distance between main and subreflector surfaces (measured along the principal ray),

$$l_{sm} = \frac{2F}{1 + \cos(\theta_0)} - \frac{(e - 1/e)f_{12}}{2[e\cos(\beta - \theta_0) - 1]} \quad (12)$$

Feed clearance, $d_{cf} = \pm[2F\tan(\frac{\theta_0}{2}) - f_{12}\sin(\beta)] - \frac{D}{2}$

Subreflector clearance,
$$d_{cs} = -[2F\tan(\frac{\theta_0}{2}) - f_{12}\sin(\beta)] - \frac{D}{2} - \frac{f_{12}(e - 1/e)\sin(\theta_p \pm \theta_e)}{2[e\cos(\alpha \pm \theta_e) + 1]}$$
$$(13)$$

In the last two equations, the positive (negative) signs are for the front (side) fed configurations. Note that these clearances must be positive and $|\alpha| > \theta_e$ for a blockage-free configuration

The front (side) fed configurations are defined as followed. Normally, two blockage-free configurations are possible by the proper choice of $\theta_p$: the "front-fed" and the "side-fed" configurations are produced by $\theta_p$ ~180°, and $\theta_p$ ~-90°, respectively.

Now, using the new optimization technique, we take the three parameters $D$, $\theta_e$, $\theta_p$ as the fixed, given primary values and vary the two parameters $f_{12}$, $\theta_0$ in order to achieve an optimum design which is compact. This is achieved by minimizing $l_{sm}$, and with no feed or subreflector blockage, namely, $d_{cf} < d_{cf0}$, and $d_{cs} < d_{cs0}$, in which $d_{cf0} \geq 0$, and $d_{cs0} \geq 0$, are given values. Thus, with a fitness function to be minimized, defined as

*Fitness = $L_{sm}$+($d_{cf}$-$d_{cf0}$)+ ($d_{cs}$-$d_{cs0}$),*    (14)

*if $L_{sm}$,($d_{cf}$-$d_{cf0}$),and ($d_{cs}$-$d_{cs0}$) are all positive, with $d_{cf0}$=1, $d_{cs0}$=1*

*Otherwise, Fitness=1000 (a relatively large number)*

We start with $D$=100, $\theta_e$=30, $\theta_p$= -170°, and upon optimization end up with the values listed in the table below

| $\theta_0$ | $f_{12}$=2c | $L_{sm}$ | $d_{cf}$ | $d_{cs}$ | $F$ | $M$ |
|---|---|---|---|---|---|---|
| -81.67 | 94.799 | 93.52 | 34.927 | 1 | 100.93 | 2.3970 |

A plot of the optimized reflector antenna geometry is given in Figure 6.

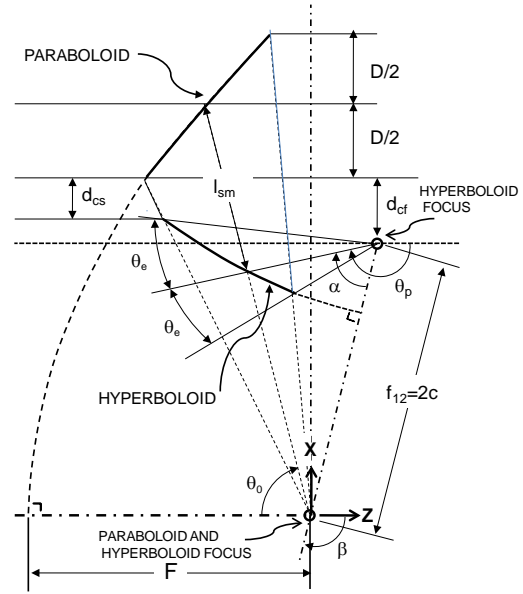

Figure 5: Geometry of the Dragonian dual reflector system.

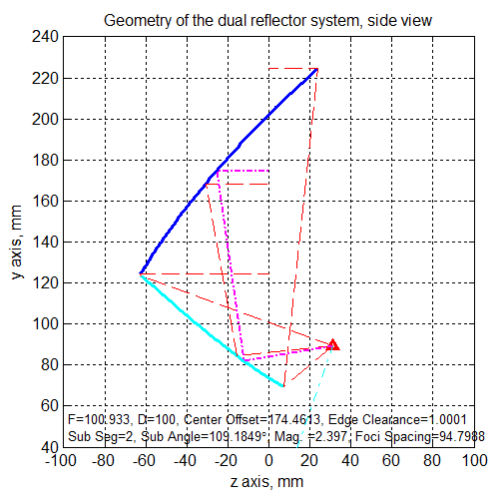

Figure 6: Geometry of optimized Dragonian system

## 5. CONCLUSIONS

We have demonstrated the properties and application of a new evolutionary optimization algorithm based on Tukey's symmetric lambda distribution. We showed that various other distributions can be approximated by this distribution, and as a result can encompass a large class of probability distributions. In addition, Because of these attributes, an Evolutionary Programming (EP) algorithm with Tukey mutation operator may perform well in a large class of optimization problems. Various schemes in implementation of EP with Tukey distribution were discussed, and the resulting algorithms were applied to selected test functions. Finally, application to a specific antenna design problem, that of optimizing a Dragonian dual reflector system was successfully demonstrated.

### ACKNOWLEDGMENT

The research described in this paper was partially carried out at the Jet Propulsion Laboratory, California Institute of Technology, under a contract with the National Aeronautics and Space Administration.